\documentclass[10pt,twocolumn,letterpaper]{article}

\usepackage[utf8]{inputenc} 
\usepackage[T1]{fontenc}    
\usepackage{booktabs}       
\usepackage{amsfonts}       
\usepackage{nicefrac}       
\usepackage{microtype}      
\usepackage{graphicx}
\usepackage{multirow}

\usepackage{tabulary}
\usepackage{colortbl}
\usepackage{makecell}
\usepackage{algorithm}
\usepackage{algpseudocode}
\usepackage{amsmath}
\usepackage{listings}
\usepackage{bm}
\usepackage{wrapfig}   

\usepackage{iccv}              

\definecolor{darkgray}{gray}{0.2}
\definecolor{lightgray}{gray}{0.8}

\definecolor{baselinecolor}{gray}{.9}
\newcommand{\baseline}[1]{\cellcolor{baselinecolor}{#1}}

\newcommand{\tablestyle}[2]{\setlength{\tabcolsep}{#1}\renewcommand{\arraystretch}{#2}\centering\footnotesize}

\newlength\savedwidth

\newlength\savewidth
\newcommand\shline{\noalign{\global\savewidth\arrayrulewidth
                            \global\arrayrulewidth 1pt}
                   \hline
                   \noalign{\global\arrayrulewidth\savewidth}}

\definecolor{iccvblue}{rgb}{0.21,0.49,0.74}
\usepackage[pagebackref,breaklinks,colorlinks,allcolors=iccvblue]{hyperref}

\def\paperID{3409} 
\def\confName{ICCV}
\def\confYear{2025}

\title{General Compression Framework for Efficient Transformer Object Tracking}

\author{Lingyi Hong\textsuperscript{\rm 1} ~
        Jinglun Li\textsuperscript{\rm 2} ~ 
        Xinyu Zhou\textsuperscript{\rm 1} ~
        Shilin Yan\textsuperscript{\rm 1} ~
        Pinxue Guo\textsuperscript{\rm 2} ~
        Kaixun Jiang\textsuperscript{\rm 2} ~
        Zhaoyu Chen\textsuperscript{\rm 2} ~ 
        \\
        Shuyong Gao\textsuperscript{\rm 1} ~
        Runze Li\textsuperscript{\rm 3} ~ 
        Xingdong Sheng\textsuperscript{\rm 3} ~
        Wei Zhang\textsuperscript{\rm 1}\footnotemark[1] ~
        Hong Lu\textsuperscript{\rm 1}\footnotemark[1] ~  
        Wenqiang Zhang\textsuperscript{\rm 1,2}\thanks{Corresponding Author}\footnotemark[1]  ~
        \\
        \textsuperscript{\rm 1}~Shanghai Key Lab of Intelligent Information Processing, \\ College of Computer Science and Artificial Intelligence, Fudan University\\
        \textsuperscript{\rm 2}~College of Intelligent Robotics and Advanced Manufacturing, Fudan University\\
        \textsuperscript{\rm 4}~Lenovo Research \\
        {\tt\small honglyhly@gmail.com, wqzhang@fudan.edu.cn} \\
}

\begin{document}
\maketitle

\begin{abstract}
  Previous works have attempted to improve tracking efficiency through lightweight architecture design or knowledge distillation from teacher models to compact student trackers. However, these solutions often sacrifice accuracy for speed to a great extent, and also have the problems of complex training process and structural limitations. Thus, we propose a general model compression framework for efficient transformer object tracking, named CompressTracker, to reduce model size while preserving tracking accuracy. Our approach features a novel stage division strategy that segments the transformer layers of the teacher model into distinct stages to break the limitation of model structure. Additionally, we also design a unique replacement training technique that randomly substitutes specific stages in the student model with those from the teacher model, as opposed to training the student model in isolation. Replacement training enhances the student model's ability to replicate the teacher model's behavior and simplifies the training process. To further forcing student model to emulate teacher model, we incorporate prediction guidance and stage-wise feature mimicking to provide additional supervision during the teacher model's compression process. CompressTracker is structurally agnostic, making it compatible with any transformer architecture. We conduct a series of experiment to verify the effectiveness and generalizability of our CompressTracker. Our CompressTracker-SUTrack, compressed from SUTrack, retains about $\mathbf{99\%}$ performance on LaSOT ($\mathbf{72.2\%}$ AUC) while achieves $\mathbf{2.42\times}$ speed up. 
  Code is available at \href{https://github.com/LingyiHongfd/CompressTracker}{here}.
\end{abstract}

\section{Introduction}

\begin{figure*}[htbp]
  \centering
  \includegraphics[width=\linewidth]{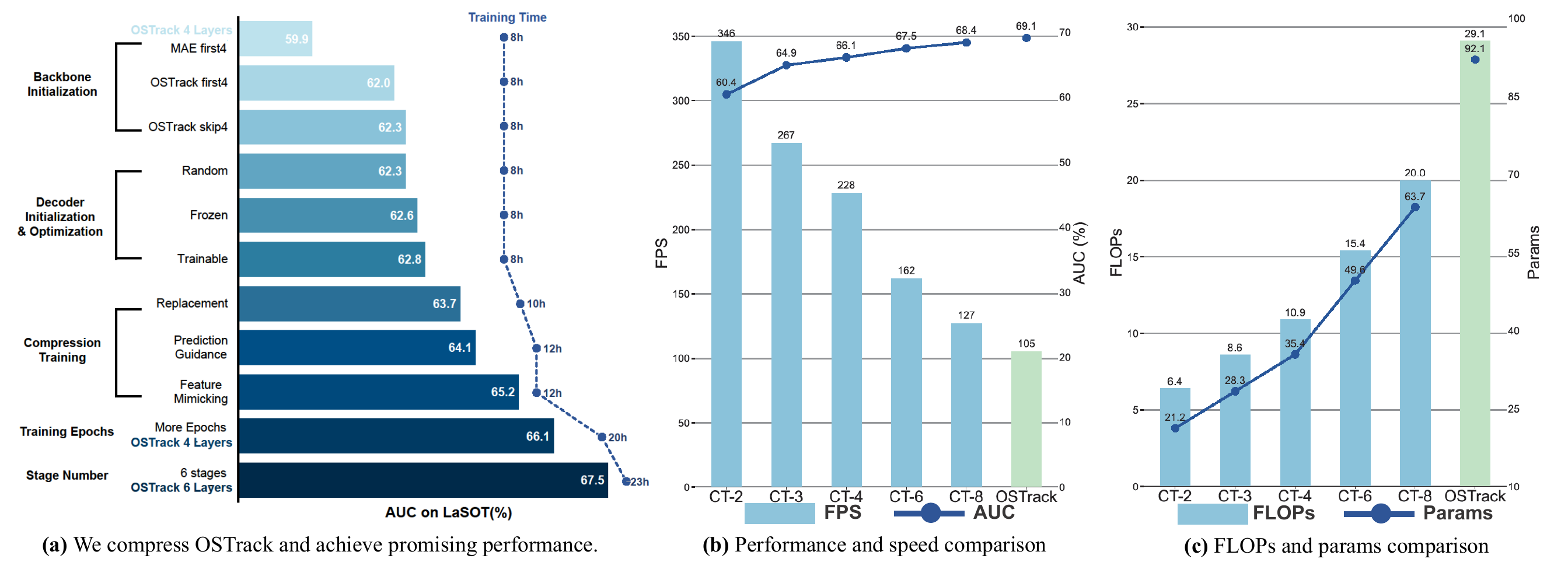}
  \caption{We apply our framework to OSTrack under several different layer configurations. \textbf{(a)} We implement each enhancement into our CompressTracker step by step. The training time is calculated by using 8 NVIDIA RTX 3090 GPUs. Notably, our CompressTracker-4 accelerates OSTrack by $\mathbf{2.17\times}$ while preserving approximately $\mathbf{96\%}$ of its original accuracy, thereby demonstrating the effectiveness of our framework. \textbf{(b)} Performance and speed comparison of CompressTracker variants with different numbers of layers. CT-x refers to a version of CompressTracker with 'x' layers. \textbf{(c)} FLOPs and parameters comparison of CompressTracker variants with different numbers of layers.}
  \label{fig:plot_all}
\end{figure*}

Visual object tracking is tasked with continuously localizing a target object across video frames based on the initial bounding box in the first frame. Transformer-based trackers have achieved promising performance on well-established benchmarks, their deployment on resource-restricted device remains a significant challenge. Developing a strong tracker with high efficiency is of great significance.

Previous works attempt to reduce model inference cost through lightweight tracker design or knowledge transfer from teacher to student models. However, these approaches face several limitations. (1) \textbf{Inferior Accuracy.} Certain works propose lightweight tracking models~\citep{borsuk2022fear,chen2022efficient,blatter2023efficient,gopal2024separable,kang2023exploring} or employ neural architecture search (NAS)~\citep{yan2021lighttrack} often suffer from underfitting and inferior performance because of the limited number of parameters. (2) \textbf{Complex Training.} Some works~\citep{cui2024mixformerv2} transfers the knowledge from a teacher tracker to a student model, while effective, requiring complex multi-stage and time-consuming training procedures and resulting in suboptimal performance. (3) \textbf{Structure Limitation.} The model reduction paradigm in~\citep{cui2024mixformerv2} severely restricts the structure of student to be consistent with teacher. 

Thus, we introduce CompressTracker, a novel and general model compression framework to enhance the efficiency of transformer tracking models, which is built upon three interconnected innovations. 
First, to address the structure limitation, we introduce the \textbf{stage division} strategy. Current dominant trackers are one-stream models~\citep{ye2022joint,cui2024mixformerv2,blatter2023efficient,chen2022efficient} with sequential transformer encoder layers that refine temporal matching features. Given this layer-wise refinement, it becomes a natural progression to consider the model not as a single entity but as a series of interconnected stages and encourage student tracker to align teacher model at each stage. Stage division partitions the teacher model into distinct stages that correspond to the layers of a simpler student model. This is achieved by dividing the teacher model into a number of stages equivalent to the student model's layers. Each stage in the student modellearns to replicate the functional behavior of its corresponding teacher stage. This division is not merely a structural alteration but a strategic educational approach, which allows student to grasp not only what to track but also how, facilitating a more effective and fine-grained compression process.

Secondly, the \textbf{stage division} lays the foundation for our \textbf{replacement training}, which tackles the complex training challenge by dynamically integrating teacher and student models during training. Replacement training strategically intertwines the teacher and student models. The core of this methodology is the dynamic substitution of stages during training. we randomly select stages from the student model and replace them with the corresponding stages from the teacher model. Replacement training permits the unaltered stages of the teacher model to collaboratively inform and enhance the learning of the substituted stages in the student model rather than supervising the entire student model as a single entity. The student model is not merely learning in parallel but is directly engaging with the teacher's learned behaviors. After training, we can just combine each stage of student model for inference. The replacement training leads to a more authentic replication of the teacher's tracking strategies and helps to prevent the student model from overfitting to specific stages of the teacher model.

Thirdly, to further improve accuracy, we introduce \textbf{prediction guidance} and \textbf{stage-wise feature mimicking}. By using the predictions of the teacher model as a reference, the student model can converge more quickly. Feature mimicking systematically aligns the feature representations learned at each stage of the student model with those of the teacher model, thereby promoting a more accurate and consistent learning. Each of these three innovations is sequentially built upon the previous one, forming a cohesive and highly effective compression framework that maximizes efficiency while preserving performance. In Figure~\ref{fig:plot_all} (a), we show procedure and results we are able to achieve with each step toward an efficient transformer tracker.

Compared to previous works, our CompressTracker holds many merits. (1) \textbf{Enhanced Mimicking and Performance.} CompressTracker enables the student model to better mimic the teacher model, resulting in better performance. 
As shown in Figure~\ref{fig:plot_all}, our CompressTracker achieves an optimal balance between accuracy and inference speed.
(2) \textbf{Simplified Training Process.} Our CompressTracker streamlines training into a single but efficient step. This simplification not only reduces the time and resources required for training but also minimizes the potential for sub-optimal performance associated with complex procedures. The training process for CompressTracker-4 requires merely 20 hours on 8 NVIDIA RTX 3090 GPUs.
(3) \textbf{Heterogeneous Model Compression.} Our stage division strategy gives a high degree of flexibility in the design of the student model. Our framework supports any transformer architecture for student model, which is not restricted to the same structure of teacher tracker. The number of layers and their structure are not predetermined but can be tailored to fit the specific computational constraints and requirements of the deployment environment. 

\begin{figure*}[htp]
  \centering
  \includegraphics[width=0.99\linewidth]{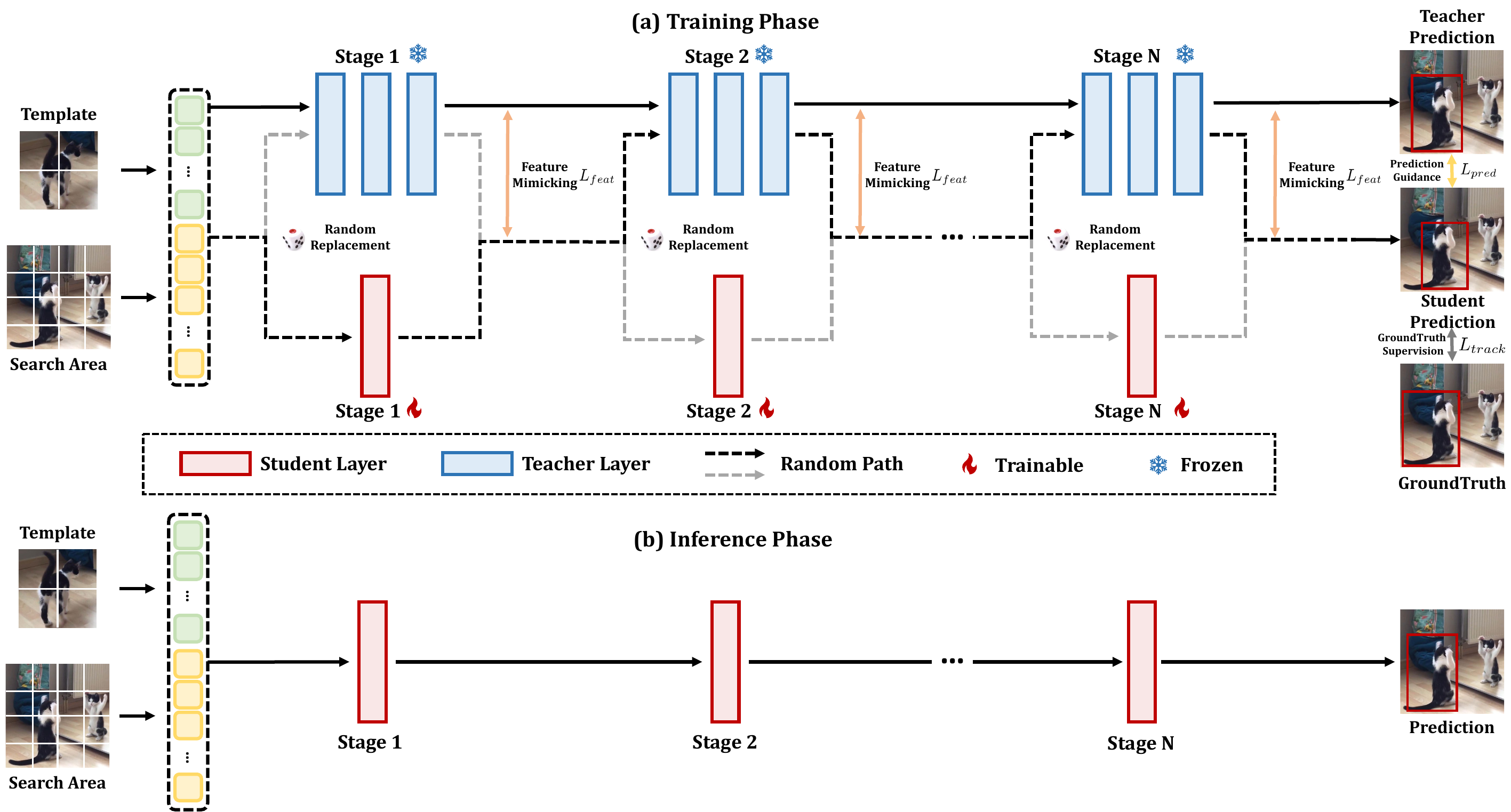}
  \caption{\textbf{CompressTracker Framework}. (a) In the training phase, we divide both the teacher model and student model into an identical number of stages. We implement a series of training strategies including replacement training, prediction guidance, and stage-wise feature mimicking, to enhance the student model's ability to emulate the teacher model. The dotted lines represent the randomly selected paths for replacement training, with black dotted lines indicating the chosen path, while gray dotted lines denote paths not selected in a specific training iteration. (b) During inference process, we simply combine each stage of the student model for testing purposes.}
  \label{fig:model}
\end{figure*}

Our contribution can be summarized as follows: (1) We introduce a novel and general model compression framework, CompressTracker, to facilitate the efficient transformer-based object tracking. (2) We propose a stage division strategy that enables a fine-grained imitation of the teacher model at the stage level, enhancing the precision and efficiency of knowledge transfer. (3) We propose the replacement training to improve the student model's capacity to replicate the teacher model's behavior. (4) We further incorporate the prediction guidance and feature mimicking to accelerate and refine the learning process of the student model. (5) Our CompressTracker breaks structural limitations, adapting to various transformer architectures for student model. Our CompressTracker-SUTrack outperforms existing models, notably accelerating SUTrack~\citep{ye2022joint} by $\mathbf{2.42\times}$ while preserving approximately $\mathbf{99\%}$ accuracy ({$\mathbf{72.2}\%$ AUC on LaSOT). 
\section{Related Work}

\textbf{Visual Object Tracking.} Visual object tracking aims to localize the target object of each frame based on its initial appearance. Previous tracking methods~\citep{bertinetto2016fully,li2018high,zhang2020ocean,danelljan2019atom,li2019siamrpn++,bolme2010visual,henriques2014high,chen2021transformer,yan2021learning,hong2024lvos,hong2023lvos} utilize a two-stream pipeline to decouple the feature extraction and relation modeling. Recently, the one-stream pipeline hold the dominant role. ~\citep{ye2022joint,cui2022mixformer,cui2024mixformerv2,bai2023artrackv2,wei2023autoregressive,chen2022backbone,chen2023seqtrack,hong2025progressive,gao2023generalized,zhou2023reading,hong2024onetracker,hong2023simulflow,zhou2024detrack} combine feature extraction and relation modeling into a unified process. These models are built upon vision transformer, which consists of a series of transformer encoder layers. Thanks to a more adequate relationship modeling between template and search frame, one-stream models achieve impressive performance. However, these models suffer from low inference efficiency, which is the main obstacle to practical deployment.

\textbf{Efficient Tracking.} 
Some works have attempted to speed up tracking models. ~\citep{yan2021lighttrack} utilizes neural architecture search (NAS) to search a light Siamese network, and the searching process is complex. ~\citep{borsuk2022fear,chen2022efficient,blatter2023efficient,kang2023exploring} design a lightweight tracking model, but the small number of parameters restricts the accuracy to a large degree. MixFormerV2~\citep{cui2024mixformerv2} propose a complex multi-stage model reduction strategy. Although MixFormerV2-S achieves real-time speed on CPU, the multi-stage training strategy is time consuming, which requires about 120 hours (5 days) on 8 Nvidia RTX8000 GPUs, even several times the original training time of MixFormer~\citep{cui2022mixformer}. Any suboptimal performance during these stages impact the final model's performance negatively. Besides, the reduction paradigm imposes constraints on the design of student models. To address these shortcuts, we propose the general model compression framework, CompressTracker, to explore the roadmap toward an end-to-end and traininig-efficient model compression for lightweight transformer-based tracker. Our CompressTracker break the structure restriction and achieves balance between speed and accuracy.

\textbf{Transformer Compression.} Model compression aims to reduce the size and computational cost of a large model while retaining as much performance as possible, and recently many attempts have been made to speed up a large pretrained transformer model. ~\citep{frankle2018lottery} reduced the number of parameters through pruning technique, and~\citep{shen2020q} accomplished the quantization of BERT to 2-bits utilizing Hessian information.~\citep{sanh2019distilbert,sun2019patient,jiao2019tinybert,xu2020bert} leverage the knowledge distillation to transfer the knowledge from teacher to student model and exploit pretrained model. Beyond language models, considerable focus has also been placed on compressing vision transformer models. ~\citep{rao2021dynamicvit,xu2022evo,chen2021autoformer,gong2022nasvit,chavan2022vision,yang2022vitkd,zhang2022minivit} utilize multiple model compression techniques to compress vision transformer models. MixFormerV2~\citep{cui2024mixformerv2} proposed a two-stage model reduction paradigm to distill a lightweight tracker, relying on the complex multi-stage distillation training. However, our CompressTracker propose an end-to-end and efficient compression training to achieve any transformer structure compression, which speed up SUTrack $\mathbf{2.42\times}$ while maintaining about $\mathbf{99\%}$ accuracy.

\section{CompressTracker}

In this section, we will introduce our proposed general model compression framework, CompressTracker. The workflow of our CompressTracker in illustrated in Figure~\ref{fig:model}.

\subsection{Stage Division}
\label{sec:method_stage}
Recently, transformer-based one-stream tracking models~\citep{chen2022backbone,cui2022mixformer,ye2022joint,cui2024mixformerv2} have dominated visual object tracking, consisting of multiple transformer encoder layers that progressively refine temporal matching features. Building upon this layer-wise refinement, we introduce the stage division strategy, which segments the model into a series of sequential stages. This approach encourages the student model to emulate the teacher model's behavior at each individual stage. Specifically, we denote the pretrained tracker and the compressed model as \textit{teacher} and \textit{student} model, with $N_{t}$ and $N$ layers, respectively. Both teacher and student models are then divided into $N$ stages, where each student stage encompasses a single layer, and each corresponding stage in teacher may aggregate multiple layers, which can be formulated as:
\begin{align}
\setlength{\abovedisplayshortskip}{-4mm}
\setlength{\belowdisplayshortskip}{-4mm}
    teacher & = \{stage_{1}^{t},stage_{2}^{t},...,stage_{N}^{t}\}, \\
    student & = \{stage_{1}^{s},stage_{2}^{s},...,stage_{N}^{s}\},
\end{align}

\noindent where $stage_{i}^{t}$ and $stage_{i}^{s}$ denote
the corresponding stage $i$ in teacher and student model, respectively.
For a specific stage $i$, we establish a correspondence between the stages of the teacher and student models. The objective of stage division is to enforce each stage of the student model to replicate its counterpart in the teacher model. This stage division strategy breaks the traditional approach that treats the model as an indivisible whole~\citep{borsuk2022fear,chen2022efficient,blatter2023efficient,cui2024mixformerv2}. Instead, it enables a fine-grained learning process where the student model transfers knowledge from the teacher in a more detailed, stage-specific manner.

Unlike the reduction paradigm adopted in \citep{cui2024mixformerv2}, our CompressTracker framework facilitates support for arbitrary transformer structures of the student tracker, thanks to our innovative stage-wise division design. To align the size and channel dimensions of the student model's temporal matching features with those of the teacher model, we implement input and output projection layers before and after the student layers, respectively. These layers ensure compatibility between teacher and student models, allowing greater flexibility in the student model’s architecture. During inference, these projection layers are omitted.

\begin{table*}[tp]
  \centering
  \resizebox{\linewidth}{!}{\begin{tabular}{l|ccc|cc|cc|ccc|cc|cc}
    \toprule
    \multirow{2}{*}{\textbf{Method}} &
    \multicolumn{3}{c|}{\textbf{LaSOT}} &
    \multicolumn{2}{c|}{\textbf{LaSOT$_{ext}$}} &
    \multicolumn{2}{c|}{\textbf{TNL2K}} &
    \multicolumn{3}{c|}{\textbf{TrackingNet}} &
    \multicolumn{2}{c|}{\textbf{UAV123}} &
    \multicolumn{2}{c}{\textbf{FPS}}  \\
    & AUC & P$_{Norm}$ & P & AUC & P & AUC & P & AUC & P$_{Norm}$ & P & AUC & P & GPU & CPU \\
    \midrule

    \textbf{CompressTracker-OSTrack\cite{ye2022joint}} & 66.1 $_{\textcolor{darkgray}{96\%}}$ & 75.2 & 70.6 & 45.7 $_{\textcolor{darkgray}{96\%}}$ & 50.8 & 53.6 $_{\textcolor{darkgray}{99\%}}$ & 52.5 & 82.1 $_{\textcolor{darkgray}{99\%}}$ & 87.6 & 80.1 & 67.4 $_{\textcolor{darkgray}{99\%}}$ & 88.0 & \textbf{228} $_{\textcolor{darkgray}{2.17\times}}$ & 18 $_{\textcolor{darkgray}{2.11\times}}$\\
    \textbf{CompressTracker-SeqTrack~\cite{chen2023seqtrack}} & 68.1 $_{\textcolor{darkgray}{95\%}}$ & 77.3 & 73.4 & 47.9 $_{\textcolor{darkgray}{96\%}}$ & 56.4 & 54.5 $_{\textcolor{darkgray}{99\%}}$ & 54.2 & 83.1 $_{\textcolor{darkgray}{98\%}}$ & 88.1 & 81.7 & 68.4 $_{\textcolor{darkgray}{98\%}}$ & 90.1 & 78 $_{\textcolor{darkgray}{1.95\times}}$ & 10 $_{\textcolor{darkgray}{2.46\times}}$ \\
    \textbf{CompressTracker-ODTrack~\cite{zheng2024odtrack}} & 70.5 $_{\textcolor{darkgray}{96\%}}$ & 79.5 & 76.9 & 50.9 $_{\textcolor{darkgray}{97\%}}$ & 60.8 & 58.2 $_{\textcolor{darkgray}{99\%}}$ & 58.9 & 82.8 $_{\textcolor{darkgray}{97\%}}$ & 87.9 & 81.4 & 69.2 $_{\textcolor{darkgray}{98\%}}$ & 90.8 & 87 $_{\textcolor{darkgray}{2.71\times}}$ & 13 $_{\textcolor{darkgray}{3.18\times}}$ \\
    \textbf{CompressTracker-SUTrack~\cite{chen2024sutrack}} & \textbf{72.2} $_{\textcolor{darkgray}{99\%}}$ & \textbf{80.7} & \textbf{78.2} & \textbf{52.1} $_{\textcolor{darkgray}{98\%}}$ & \textbf{63.2} & \textbf{64.4} $_{\textcolor{darkgray}{99\%}}$ & \textbf{69.9} & \textbf{84.3} $_{\textcolor{darkgray}{99\%}}$ & \textbf{88.2} & \textbf{83.3} & \textbf{71.0} $_{\textcolor{darkgray}{99\%}}$ & \textbf{91.4} & 134 $_{\textcolor{darkgray}{2.42\times}}$ & \textbf{24} $_{\textcolor{darkgray}{2.34\times}}$\\

    \midrule

    SUTrack-T~\cite{chen2024sutrack} & 69.6 & 79.3 & 75.4 & 50.2 & 57.0 & - & - & 82.7 & 87.2 & 80.8 & 69.4 & - & 100 & 23\\
    
    HiT-Base~\citep{kang2023exploring} & 64.6 & 73.3 & 68.1 & 44.1 & - & - & - & 80.0 & 84.4 & 77.3 & 65.6 & - & 175 & 33 \\
    HiT-Samll~\citep{kang2023exploring} & 60.5 & 68.3 & 61.5 & 40.4 & - & - & - & 77.7 & 81.9 & 73.1 & 63.3 & - & 192 & 72\\
    HiT-Tiny~\citep{kang2023exploring} & 54.8 & 60.5 & 52.9 & 35.8 & - & - & - & 74.6 & 78.1 & 68.8 & 53.2 & - & 204 & 76 \\

    SMAT~\citep{gopal2024separable} & 61.7 & 71.1 & 64.6 & - & - & - & - & 78.6 & 84.2 & 75.6 & 64.3 & 83.9 & 158 & 33 \\
    
    MixFormerV2-S~\citep{cui2024mixformerv2} & 60.6 & 69.9 & 60.4 & 43.6 & 46.2 & 48.3 & 43.0 & 75.8 & 81.1 & 70.4 & 65.8 & 86.8 & 325 & 30\\
   
    FEAR-L~\citep{borsuk2022fear} & 57.9 & 68.6 & 60.9 & - & - & - & - & - & - & - & - & - & - & - \\ 
    FEAR-XS~\citep{borsuk2022fear} & 53.5 & 64.1 & 54.5 & - & - & - & -  & - & - & - & - & - & 80 & 60 \\ 
    HCAT~\citep{chen2022efficient} & 59.0 & 68.3 & 60.5 & - & - & - & - & 76.6 & 82.6 & 72.9 & 63.6 & - & 195 & 45 \\
    E.T.Track~\citep{blatter2023efficient} & 59.1 & - & - & - & - & - & - & 74.5 & 80.3 & 70.6 & 62.3 & - & 150 & 47\\ 
    LightTrack-LargeA~\citep{yan2021lighttrack} & 55.5 & - & 56.1 & - & - & - & - & 73.6 & 78.8 & 70.0 & - & - & - & - \\
    LightTrack-Mobile~\citep{yan2021lighttrack} & 53.8 & - & 53.7 & - & - & - & - & 72.5 & 77.9 & 69.5 & -  & - & 120 & 41 \\
    STARK-Lightning~\citep{yan2021learning} & 58.6 & 69.0 & 57.9 & - & -  & - & - & - & - & - & - & - & 200 & -\\
    DiMP~\citep{bhat2019learning} & 56.9 & 65.0 & 56.7 & - & - & - & - & 74.0 & 80.1 & 68.7 & 65.4 & - & 77 & - \\
    SiamFC++~\citep{xu2020siamfc++} & 54.4& 62.3 & 54.7 & - & - & - & - & 75.4 & 80.0 & 70.5 & - & -  & 90 & - \\

    \bottomrule
    \end{tabular}}
  \caption{\textbf{Compress Object Tracker.} We compress several trackers such as SUTrack~\cite{chen2024sutrack}, ODTrack~\cite{zheng2024odtrack}, and so on, into students with an optimal trade-off between accuracy and efficiency. We report the performance on 5 benchmarks and calculate the performance gap in comparison to the origin teacher model. CompressTracker outperforms previous models in both speed and accuracy. }
  \label{tab:main_all}
\end{table*}

\subsection{Replacement Training}
\label{sec:repalce_training}
During training, we adopt the replacement training to integrates teacher model and student models, diverging from the conventional practice of training the student model in isolation. In each iteration, a stochastic process determines which student model stages are replaced by corresponding teacher stages. For the specific stage $i$, the forward process in conventional isolated propagation can be described as:

\begin{equation}
    h_{i} = stage_{i}^{s}(h_{i-1}),
\end{equation}

\noindent where $h_{i-1}$ is the input of the $i$ student stage. 
However, in our replacement training, we decide whether to replace or not by random Bernoulli sampling $b_{i}$ with probability $p$, where $b_{i} \in \{0,1\}$. If $b_{i}$ equals $1$, the output from the preceding $\mathit{i-1}$-th stage is directed to the $i$ student stage, otherwise, we channel the output into the $i$ frozen teacher stage, which can be formulated as:

\begin{equation}
\setlength{\abovedisplayshortskip}{-3mm}
\setlength{\belowdisplayshortskip}{-1mm}
    h_{i} = \left\{ \begin{array}{ll}
        stage_{i}^{t}(h_{i-1}),   & r_{i} =0, \\
        stage_{i}^{s}(h_{i-1}),   & r_{i} =1, 
        \end{array} \right. r_{i} \sim \mathrm{Bernoulli}(p).
\end{equation}

This replacement training creates a collaborative learning environment where the teacher model dynamically supervises the student model. The unreplaced stages of teacher provide valuable contextual supervision for a specific stage in the student model. Consequently, the student model is not operating in parallel but is actively engaged with and learning from the teacher's established behaviors. For the optimization of student model, we only require the groundtruth box and denote the loss as $L_{track}$. After training, the student model's stages are harmoniously combined for inference.

\subsection{Prediction Guidance \& Stage-wise Feature Mimicking}
Replacement training enables the student model to learn the behavior of each individual stage, resulting in enhanced performance. However, mimicking the teacher directly can be too difficult for a smaller student model. Thus, we employ the teacher's predictions to further guide the learning of compressed tracker. We apply the same loss as $L_{track}$ for prediction guidance, which is denoted as $L_{pred}$. This guidance accelerates and stabilizes the student's learning, helping it assimilate knowledge from the teacher more effectively.

While prediction guidance accelerates the convergence, the student tracker might not entirely match the complex behavior of the teacher model. We introduce the stage-wise feature mimicking to further synchronize the temporal matching features between corresponding stages of the teacher and student models. This alignment is quantified by calculating the $L_{2}$ distance between the outputs of these stages, which is referred as $L_{feat}$. It is worth noting that any metric assessing the discrepancy in feature distributions can serve as the loss function. However, we choose a simple $L_{2}$ distance rather than a complex loss to highlight the effectiveness of our stage division and replacement training strategies. The stage-wise feature mimicking both promotes a closer similarity in feature representations of corresponding stages and enhances overall coherence between teacher and student.

\subsection{Progressive Replacement}
\label{sec:progress}
In Section~\ref{sec:repalce_training}, we describe the replacement training strategy. Although setting the Bernoulli sampling probability $p$ as a constant value can realize the compression, these stages have not been trained together at the same time and there may be some dissonance. A further finetuning step is necessary to achieve better harmony among the stages. Thus, we introduce a progressive replacement strategy to bridges the gap between the two initially separate training phases, fostering an end-to-end easy-to-hard learning process. By adjusting the value of $p$, we can control the number of stages to be replaced. The value of  $p$ gradually increases from $p_{init}$ to $1.0$, allowing for a more incremental and coherent training:

\begin{equation}
\setlength{\abovedisplayshortskip}{-3mm}
\setlength{\belowdisplayshortskip}{-1mm}
p = \left\{ \begin{array}{ll}
        p_{init},   & 0\leq t < \alpha_{1} m, \\
        p_{init}\frac{m+t-2\alpha_{1}m-\alpha_{2}m}{(1-\alpha_{1}-\alpha_{2})m} & \alpha_{1} m \leq t \leq (1-\alpha_{2}) m, \\
        1.0,   & (1-\alpha_{2}) m < t \leq m,
        \end{array} \right.
\end{equation}

\noindent where $m$ represents the total number of training epochs, and $t$ is a specific training epoch, $\alpha_{1}$ and $\alpha_{2}$ are hyper parameters to modulate the training process. Specifically, $\alpha_{1}$ controls the duration of warmup process, whereas $\alpha_{2}$ determines the length of final finetuning process. The mathematical expectation of $p$ for each layer is:

{
\setlength{\abovedisplayshortskip}{-3mm}
\setlength{\belowdisplayshortskip}{-1mm}
\begin{equation}
E(p) = \int_{0}^{m} p dt = [\frac{1+p_{init}}{2} + \frac{1-p_{init}}{2}(\alpha_{2}-\alpha_{1})] m.
\end{equation}
}

It is worth noting that each layer is optimized fewer times than the total iteration count, according to the mathematical expectation. Through dynamically adjusting the replacement rate $p$, we eliminate the requirement of finetuning and accomplish an end-to-end model compression.

\begin{table*}[t]
  \centering
  \resizebox{\linewidth}{!}{\begin{tabular}{l|ccc|cc|cc|ccc|cc|c}
    \toprule
    \multirow{2}{*}\textbf{{Method}} &
    \multicolumn{3}{c|}{\textbf{LaSOT}} &
    \multicolumn{2}{c|}{\textbf{LaSOT$_{ext}$}} &
    \multicolumn{2}{c|}{\textbf{TNL2K}} &
    \multicolumn{3}{c|}{\textbf{TrackingNet}} &
    \multicolumn{2}{c|}{\textbf{UAV123}} &
    \multirow{2}{*}{\textbf{FPS}}  \\
    & AUC & P$_{Norm}$ & P & AUC & P & AUC & P & AUC & P$_{Norm}$ & P & AUC & P &  \\
    \midrule
    \multicolumn{14}{c}{\textit{Model Generalization}}\\
    \midrule    
    \textbf{CompressTracker-OSTrack~\cite{ye2022joint}} & 66.1 $_{\textcolor{darkgray}{96\%}}$ & 75.2 & 70.6 & 45.7 $_{\textcolor{darkgray}{96\%}}$ & 50.8 & 53.6 $_{\textcolor{darkgray}{99\%}}$ & 52.5 & 82.1 $_{\textcolor{darkgray}{99\%}}$ & 87.6 & 80.1 & 67.4 $_{\textcolor{darkgray}{99\%}}$ & 88.0 & 228 $_{\textcolor{darkgray}{2.17\times}}$ \\
    \textbf{CompressTracker-SeqTrack~\cite{chen2023seqtrack}} & 68.1 $_{\textcolor{darkgray}{95\%}}$ & 77.3 & 73.4 & 47.9 $_{\textcolor{darkgray}{96\%}}$ & 56.4 & 54.5 $_{\textcolor{darkgray}{99\%}}$ & 54.2 & 83.1 $_{\textcolor{darkgray}{98\%}}$ & 88.1 & 81.7 & 68.4 $_{\textcolor{darkgray}{98\%}}$ & 90.1 & 78 $_{\textcolor{darkgray}{1.95\times}}$ \\
    \textbf{CompressTracker-ODTrack~\cite{zheng2024odtrack}} & 70.5 $_{\textcolor{darkgray}{96\%}}$ & 79.5 & 76.9 & 50.9 $_{\textcolor{darkgray}{97\%}}$ & 60.8 & 58.2 $_{\textcolor{darkgray}{99\%}}$ & 58.9 & 82.8 $_{\textcolor{darkgray}{97\%}}$ & 87.9 & 81.4 & 69.2 $_{\textcolor{darkgray}{98\%}}$ & 90.8 & 87 $_{\textcolor{darkgray}{2.71\times}}$ \\
    \textbf{CompressTracker-SUTrack~\cite{chen2024sutrack}} & 72.2 $_{\textcolor{darkgray}{99\%}}$ & 80.7 & 78.2 & 52.1 $_{\textcolor{darkgray}{98\%}}$ & 63.2 & 64.4 $_{\textcolor{darkgray}{99\%}}$ & 69.9 & 84.3 $_{\textcolor{darkgray}{99\%}}$ & 88.2 & 83.3 & 71.0 $_{\textcolor{darkgray}{99\%}}$ & 91.4 & 134 $_{\textcolor{darkgray}{2.42\times}}$ \\
    \midrule
    \multicolumn{14}{c}{\textit{Stage Scalability}}\\
    \midrule
    \textbf{CompressTracker-OSTrack-2} & 60.4 $_{\textcolor{darkgray}{87\%}}$ & 68.5 & 61.5 & 40.4 $_{\textcolor{darkgray}{85\%}}$ & 43.8 & 48.5 $_{\textcolor{darkgray}{89\%}}$ & 45.0 & 78.2 $_{\textcolor{darkgray}{94\%}}$ & 83.3 & 74.8 & 62.5  $_{\textcolor{darkgray}{92\%}}$ & 82.5 & 346 $_{\textcolor{darkgray}{3.30\times}}$ \\
    \textbf{CompressTracker-OSTrack-3} & 64.9 $_{\textcolor{darkgray}{94\%}}$ & 74.0 & 68.4 & 44.6 $_{\textcolor{darkgray}{94\%}}$ & 49.6 & 52.6 $_{\textcolor{darkgray}{97\%}}$ & 50.9 & 81.6 $_{\textcolor{darkgray}{98\%}}$ & 86.7 & 79.4 & 65.4 $_{\textcolor{darkgray}{96\%}}$ & 88.3 & 267 $_{\textcolor{darkgray}{2.54\times}}$ \\
    \textbf{CompressTracker-OSTrack-4} & 66.1 $_{\textcolor{darkgray}{96\%}}$ & 75.2 & 70.6 & 45.7 $_{\textcolor{darkgray}{96\%}}$ & 50.8 & 53.6 $_{\textcolor{darkgray}{99\%}}$ & 52.5 & 82.1 $_{\textcolor{darkgray}{99\%}}$ & 87.6 & 80.1 & 67.4 $_{\textcolor{darkgray}{99\%}}$ & 88.0 & 228 $_{\textcolor{darkgray}{2.17\times}}$ \\
    \textbf{CompressTracker-OSTrack-6} & 67.5 $_{\textcolor{darkgray}{98\%}}$ & 77.5 & 72.4 & 46.7 $_{\textcolor{darkgray}{99\%}}$ & 52.5 & 54.7 $_{\textcolor{darkgray}{101\%}}$ & 54.3 & 82.9 $_{\textcolor{darkgray}{99\%}}$ & 87.8 & 81.5 & 67.9 $_{\textcolor{darkgray}{99\%}}$ & 88.7 & 162 $_{\textcolor{darkgray}{1.54\times}}$ \\
    \textbf{CompressTracker-OSTrack-8} & 68.4 $_{\textcolor{darkgray}{99\%}}$ & 78.0 & 73.1 & 47.2 $_{\textcolor{darkgray}{99\%}}$ & 53.1 & 55.2 $_{\textcolor{darkgray}{102\%}}$ & 54.8 & 83.3 $_{\textcolor{darkgray}{101\%}}$ & 88.0 & 81.9 & 68.2 $_{\textcolor{darkgray}{99\%}}$ & 89.0 & 127 $_{\textcolor{darkgray}{1.21\times}}$ \\
    \midrule
    \multicolumn{14}{c}{\textit{Larger Transformer Scalability}}\\
    \midrule
    \textbf{CompressTracker-OSTrack-L-4} & 67.5 $_{\textcolor{darkgray}{96\%}}$ & 77.9 & 72.7 & 45.9 $_{\textcolor{darkgray}{98\%}}$ & 51.5 & 58.3 $_{\textcolor{darkgray}{98\%}}$ & 58.5 & 83.2 $_{\textcolor{darkgray}{99\%}}$ & 87.7 & 81.2 & 67.4 $_{\textcolor{darkgray}{99\%}}$ & 88.2 & 228 $_{\textcolor{darkgray}{2.84\times}}$ \\
    \midrule
    \multicolumn{14}{c}{\textit{Higher Resolution Scalability}}\\
    \midrule
    \textbf{CompressTracker-OSTrack-384-4} & 67.7 $_{\textcolor{darkgray}{96\%}}$ & 78.3 & 73.6 & 48.1 $_{\textcolor{darkgray}{96\%}}$ & 58.0 & 54.3 $_{\textcolor{darkgray}{99\%}}$ & 53.7 & 82.7 $_{\textcolor{darkgray}{99\%}}$ & 87.3 & 81.0 & 68.2 $_{\textcolor{darkgray}{98\%}}$ & 88.8 & 228 $_{\textcolor{darkgray}{3.90\times}}$ \\

    \bottomrule
    \end{tabular}}
  \caption{\textbf{Generalization of CompressTracker.} Our CompressTracker can be applied to any teacher model, any resolution, any stage number, any teacher model size, and any student architectures. CompressTracker-m-x denotes the compressed student model with 'x' layers with 'm' as teacher model. We report the performance on 5 benchmarks and calculate the performance gap in comparison to the origin teacher model.}
  \label{tab:main_general}
\end{table*}

\subsection{Inner Connection of CompressTracker}
The contributions of our CompressTracker are sequential and deeply interconnected  for efficient transformer-based tracking. CompressTracker begins with the stage division strategy,is the foundation of our framework, as it directly enables replacement training, where student model stages are dynamically substituted with teacher model stages during training. Without this structured stage division, replacement training would not be feasible, underscoring the necessity of our initial design. Building on this, we further enhance the knowledge transfer process with prediction guidance and stage-wise feature mimicking. These components work in tandem, each reinforcing the next, ensuring that knowledge is transferred effectively at every stage. Our CompressTracker maintains a high level of efficiency throughout the compression process, which allows us to achieve an optimal balance between accuracy and efficiency, and highlights the inner connection and superiority of each contribution. 

\subsection{Training and Inference}
Our CompressTracker is a general framework applicable to various student model architectures. It simplifies the student model optimization by using a straightforward end-to-end training process, without the need for multi-stage methods or complex loss functions. During training, teacher model is frozen and we only optimize student tracker. The total loss for CompressTracker is:

\begin{equation}
L=\lambda_{track}L_{track}+\lambda_{pred}L_{pred}+\lambda_{feat}L_{feat}.
\end{equation}

After training, the various stages of the student model are combined to create a unified model for the inference phase. Consistent with previous methods~\citep{ye2022joint,cui2022mixformer}, a Hanning window penalty is adopted.

\section{Experiments}
\subsection{Implement Details}
Our framework CompressTracker is general and not dependent on a specific transformer structure, hence we select a series of trackers, such as SUTrack~\cite{chen2024sutrack}, ODTrack~\cite{zheng2024odtrack}, SeqTrack~\cite{chen2023seqtrack}, OSTrack~\citep{ye2022joint} and so on as baseline to verify the effectiveness of our CompressTracker. The training datasets consist of LaSOT~\citep{fan2019lasot}, TrackingNet~\citep{muller2018trackingnet}, GOT-10K~\citep{huang2019got}, and COCO~\citep{lin2014microsoft}, following OSTrack~\citep{ye2022joint} and MixFormerV2~\citep{cui2024mixformerv2}. We set $\lambda_{track}$ as $1$, $\lambda_{pred}$ as $1$, and $\lambda_{feat}$ as $0.2$. The $p_{init}$ is set as $0.5$. We train the CompressTracker with AdamW optimizer~\citep{loshchilov2017decoupled}, with the weight decay as $10^{-4}$ and the initial learning rate of $4 \times 10^{-5}$. The batch size is $128$. The total training epochs is 500 with 60K image pairs per epoch and the learning rate is reduced by a factor of 10 after 400 epochs. $\alpha_{1}$ and $\alpha_{2}$ are set as 0.1. The search and template images are resized to resolutions of $256\times 256$ and $128\times 128$. We initialize the CompressTracker with the pretrained parameters. We report the inference speed on a NVIDIA RTX 2080Ti GPU and Intel(R) Xeon(R) Platinum 8268 CPU @ 2.90GHz.

\subsection{Compress Object Tracker}
To validate the effectiveness of CompressTracker, we apply our CompressTracker framework to compress existing transformer-based tracking models, demonstrating its ability to balance performance and inference speed. We carefully select model configurations that offer an optimal trade-off between accuracy and efficiency. The experimental results are presented in Table~\ref{tab:main_all}.
Our method achieves significant acceleration while preserving tracking performance. For instance, when compressing SUTrack (73.2 AUC on LaSOT), our CompressTracker maintains $99\%$ performance (72.2 AUC on LaSOT) while achieving a 2.42× speedup. Furthermore, compared to existing lightweight tracking models, CompressTracker consistently outperforms them in both speed and accuracy, highlighting its superiority and strength.

\begin{table*}[t]
\centering
\begin{minipage}[t]{0.46\linewidth}
    \centering
    \tablestyle{8.5mm}{0.95}
    \begin{tabular}{ccc}
    \toprule
    \# & Init. method & AUC \\
    \shline
    1 & MAE-first4 & 59.9\%  \\
    2 & OSTrack-first4 &   62.0\%\\
    \baseline{\textbf{3}} & \baseline{\textbf{OSTrack-skip4}} & \baseline{\textbf{62.3\%}} \\
    \shline
\end{tabular}
    \vspace{-3mm}
    \captionof{table}{\textbf{Backbone Initialization.} 'MAE-first4' denotes initializing the student model using the first 4 layers of MAE-B. 'OSTrack-skip4' represents utilizing every fourth layer of OSTrack for the student model.}
    \label{tab:abl_backbone}
\end{minipage}
\hfill
\begin{minipage}[t]{0.5\linewidth}
    \centering
    \tablestyle{8.6mm}{0.95}
    \begin{tabular}{ccc}
    \toprule
    \# & Init. \& Opt. & AUC \\
    \shline
    1 & Random \& Trainable & 62.3\%  \\
    2 & Teacher \& Frozen & 62.6\%\\
    \baseline{\textbf{3}} & \baseline{\textbf{Teacher \& Trainable}} & \baseline{\textbf{62.8\%}} \\ 
    \shline
\end{tabular}
    \vspace{-3mm}
    \captionof{table}{\textbf{Decoder Initialization and Optimization.} 'Random' denotes randomly initialized decoder, and 'Teacher' means the decoder is initialized with teacher parameters. 'Frozen' represents that the decoder is frozen, and 'Trainable' denotes decoder is trainable.}
    \label{tab:abl_decoder}
    
\end{minipage}

\vspace{1mm} 

\begin{minipage}[t]{0.36\linewidth}
    \centering
    \tablestyle{6.8mm}{1.0}
    \begin{tabular}{ccc}
    \toprule
    \# & Layer Split & AUC \\
    \shline
    \baseline{\textbf{1}} & \baseline{\textbf{Even}} & \baseline{\textbf{62.8\%}}  \\
    2 & Uneven & 62.7\%\\
    \shline
\end{tabular}
    \vspace{-3mm}
    \captionof{table}{\textbf{Stage Division.} 'Even' denotes evenly dividing stage, and 'Uneven' means that the layer number of each stage in teacher model is 2,2,6,2.}
    \label{tab:abl_layer}
\end{minipage}
\hfill
\begin{minipage}[t]{0.36\linewidth}
    \centering
    \tablestyle{2.7mm}{1.0}
    \begin{tabular}{cccc}
    \toprule
    \# & Replacement & AUC & Training Time\\
    \shline
    \baseline{\textbf{1}} & \baseline{\textbf{Random}} & \baseline{\textbf{65.2\%}} & \baseline{\textbf{12 h}}  \\
    2 & Decouple-300 & 64.6\% & 16 h \\
    \shline
\end{tabular}
    \vspace{-3mm}
    \captionof{table}{\textbf{Replacement training.} 'Random' denotes our replacement training, and 'Decouple-300' represents decoupling the training of each stage.}
    \label{tab:ablation_replace}
\end{minipage}
\hfill
\begin{minipage}[t]{0.26\linewidth}
    \centering
    \tablestyle{2.8mm}{1.0}
    \begin{tabular}{ccc}
    \toprule
    \# & Replacement & AUC \\
    \shline
    \baseline{\textbf{1}} & \baseline{\textbf{w/ Progressive}} & \baseline{\textbf{65.2\%}}  \\
    2 & w/o Progressive & 64.8\% \\
    \shline
\end{tabular}
    \vspace{-3mm}
    \caption{\textbf{Progressive Replacement.} We conduct additional finetune if pregressive replacement is disabled.}
    \label{tab:ablation_replace_ori}
\end{minipage}

\vspace{1mm}

\begin{minipage}[t]{0.32\linewidth}
    \centering
    \tablestyle{6.5mm}{1.0}
    \begin{tabular}{ccc}
    \toprule
    \# & Epochs & AUC \\
    \shline
    1 & 300 & 65.2\%  \\
    \baseline{\textbf{2}} & \baseline{\textbf{500}} & \baseline{\textbf{66.1\%}} \\
    \shline
\end{tabular}
    \vspace{-3mm}
    \caption{\textbf{Training Epochs.} '300' and '500' denote total epochs.}
    \label{tab:abl_epoch}
\end{minipage}
\hfill
\begin{minipage}[t]{0.32\linewidth}
    \centering
    \tablestyle{2.4mm}{0.73}
    \begin{tabular}{ccc}
    \toprule
    \# & Model & Training Time  \\
    \shline
    \baseline{\textbf{1}} & \baseline{\textbf{CompressTracker-4}} & \baseline{\textbf{20 h}}  \\
    2 & OSTrack & 17 h\\
    3 & MixFormerV2-S & 120 h\\
    
    \shline
\end{tabular}
    \vspace{-3mm}
    \caption{\textbf{Training Time} comparison with other methods. }
    \label{tab:ablation_time}
\end{minipage}
\hfill
\begin{minipage}[t]{0.32\linewidth}
    \centering
    \tablestyle{1.7mm}{0.73}
    \begin{tabular}{lccc}
    \toprule

    \# & Method & AUC & FPS \\
    
    \shline

    1 & \baseline{\textbf{CompressTracker-4}} & \baseline{66.1} & \baseline{228} \\
    2 & Distillation               & 63.8 & 228 \\
    3 & Pruning (MixFormerV2-S)    & 60.6 & 325 \\

    \bottomrule
    \end{tabular}
    \vspace{-3mm}
    \caption{\textbf{Comparison with Other Compression Techniques.}}
    \label{tab:x_other_compress}
\end{minipage}
\caption{\textbf{Ablation studies on LaSOT.} The default choice for our model is colored in \colorbox{baselinecolor}{gray}.}
\label{tab:ablations} 
\end{table*}

\subsection{Generalization Verification}
In the previous section, we demonstrate the effectiveness of our CompressTracker. To further validate its generalization ability, we conduct additional experiments across diverse settings. The results, shown in Table~\ref{tab:main_general}, confirm that our method maintains strong performance across different model architectures, stage numbers, input resolutions, and parameter scales. We also show the results with different structure of student in Supplementary Materials. Moreover, CompressTracker is highly efficient in training, requiring only 20 hours for CompressTracker-OSTrack-4 using 8 NVIDIA RTX 3090 GPUs, in contrast to the 120 hours needed for MixFormerV2-S. This training efficiency further highlights the strong generalization capability of our method, as it can rapidly adapt to various configurations while maintaining high accuracy. Based on the extensive experiment results, our CompressTracker has both highly effectiveness and stronge generalization. CompressTracker maximizes performance retention while significantly improving inference speed and is applicable to any \textit{teacher model}, any \textit{resolution}, any \textit{stage number}, any \textit{teacher model size}, and any \textit{student architectures}, which previous methods cannot achieve.

\subsection{Ablation Study}
In this section, we conduct a series of ablation studies on LaSOT to explore the factors contributing to the effectiveness of our CompressTracker. Unless otherwise specified, the teacher model is OSTrack, and the student model has 4 encoder layers. The student model is trained for 300 epochs and  $p_{init}$ is set as 0.5. Please see Supplementary Materials for more analysis.

\textbf{Backbone Initialization.} 
We initialize the backbone of student model with different parameters and only train the student model with groundtruth supervision. The results are shown in Table~\ref{tab:abl_backbone}. It can be observed that utilizing the knowledge from teacher model is crucial. Moreover, initializing with skipped layers (\#3) yields slightly better performance than continuous layers. This suggests that initialization with skipped layers leads to improved representation similarity.

\textbf{Decoder Initialization and Optimization.} 
We investigate the influence of decoder's initialization and optimization on the accuracy of student tracker in Table~\ref{tab:abl_decoder}. Initializing the decoder with parameters from the teacher model (\#2) results in an improvement of approximately $0.3\%$ compared to a decoder initialized randomly (\#1), which underscores the benefits of transferring knowledge from the teacher model to enhance the accuracy of the student model's decoder. Furthermore, making the decoder trainable leads to an additional improvement of 0.2\%.

\textbf{Stage Division.}
Our stage division strategy divides the teacher model into the several stages, and we explore the stage division strategy in Table~\ref{tab:abl_layer}. We design two kinds of division strategy: even and uneven, For the even division, we evenly split the teacher model's 12 layers into 4 stages, with each stage comprising 3 layers. For uneven division, we follow the design manner in~\citep{he2016deep,liu2022convnet} and divide the 12 layers at a ratio of 1:1:3:1. Consequently, the number of layers in each stage of the teacher model is 2, 2, 6, and 2, respectively. The performance of the two approaches is comparable, leading us to select the equal division strategy for simplicity.

\begin{figure}[htp]
  \centering
  \includegraphics[width=\linewidth]{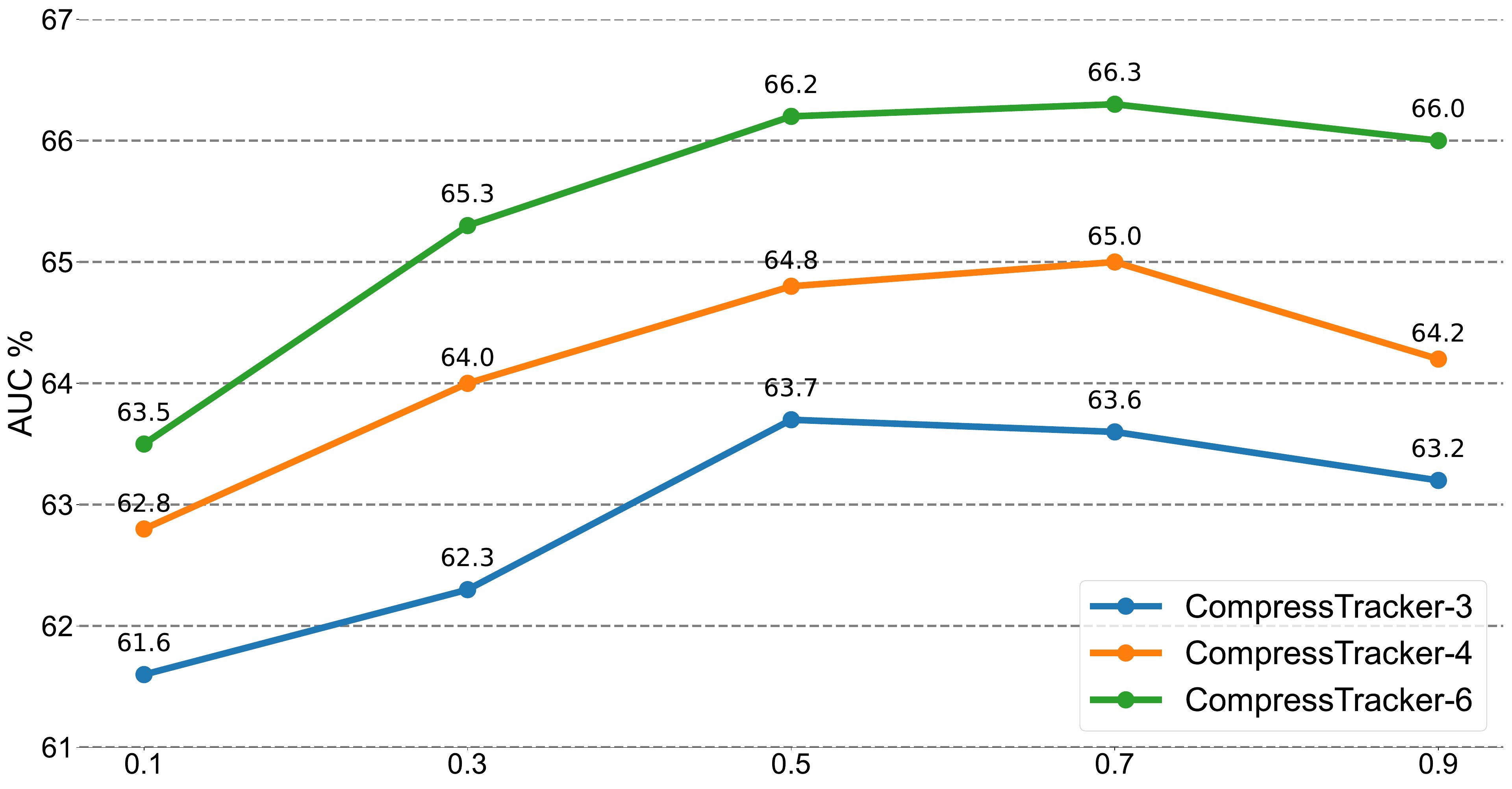}
  \caption{\textbf{Ablation study} on different replacement probability.}
  \label{fig:plot_prob}
\end{figure}

\textbf{Replacement Probability.} 
We investigate the impact of replacement probability on the accuracy of student model in Figure~\ref{fig:plot_prob}. We maintain a constant replacement probability instead of implementing the progressive replacement strategy and train the student model with 300 epochs and 30 extra finetuning epochs. It can be observed from Figure~\ref{fig:plot_prob} that performance is adversely affected when the replacement probability is set either too high or too low. Optimal results are achieved when the replacement probability is within the range of 0.5 to 0.7. Specifically, a too low probability leads to inadequate training, whereas a too high probability may result in the insufficient interaction between teacher model and student tracker. Thus, we set the $p_{init}$ as $0.5$ based on the experiment result.

\begin{figure}[htp]
  \centering
  \includegraphics[width=\linewidth]{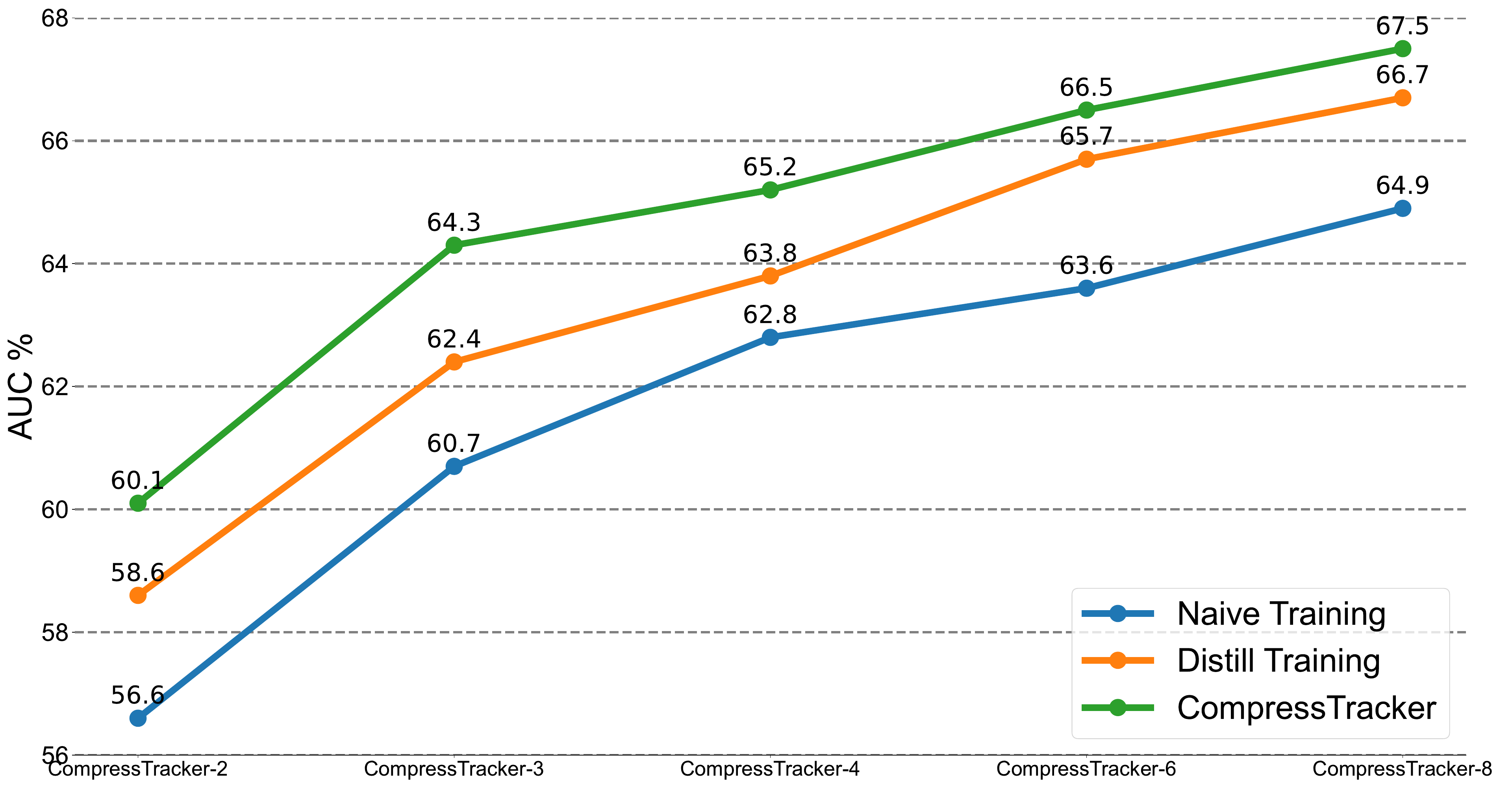}
  \caption{\textbf{Ablation study} on training strategy.}
  \label{fig:performance_plot}
\end{figure}

\begin{table}[t]
  \vspace{2mm}
  \centering
  \scriptsize
  \resizebox{\linewidth}{!}{\begin{tabular}{ccccc}
    \toprule
    \# & \makecell[c]{Prediction \\ Guidance}   & \makecell[c]{Feature \\ Mimicking} & \makecell[c]{Replacement \\ Traininig} & AUC \\
    \midrule
    1 &             &             &             &  62.8    \\
    2 & \checkmark  &             &             &  63.5    \\
    3 &             & \checkmark  &             &  63.3    \\
    4 &             &             & \checkmark  &  63.7    \\
    5 & \checkmark  &             & \checkmark  &  64.1    \\
    6 &             & \checkmark  & \checkmark  &  64.5    \\
    7 & \checkmark  & \checkmark  &             &  64.3    \\
    \baseline{\textbf{8}} & \baseline{\checkmark}  & \baseline{\checkmark}  & \baseline{\checkmark}  &  \baseline{\textbf{65.2}}    \\
    \bottomrule
  \end{tabular}}
  \caption{\textbf{Ablation study} on LaSOT about supervision of student model. The default choice for our model is colored in \colorbox{baselinecolor}{gray}.}
  \label{tab:abl_supervision}
\end{table}

\textbf{Analysis on Supervision.}
We conduct a series of experiments to comprehensively analyze the supervision effects on the student model and to verify the effectiveness of our proposed training strategy. Results are presented in Table~\ref{tab:abl_supervision}. Our proposed replacement training approach (\#4) improves by 0.9 \% AUC compared to singly training student model on groundtruth (\#1), which demonstrates that the replacement training enhances the similarity between teacher and student models. Besides, prediction guidance (\#5) and feature mimicking (\#8) further boost the performance, indicating the effectiveness of the two strategies. Compared to only training on groundtruth (\#1), our proposed replacement training, prediction guidance and feature mimicking collectively assist student model in more closely mimicking the teacher model, resulting in a total increase of $2.4 \%$ AUC.

To further explore the generalization ability of our proposed training strategy, we compare the performance of models with different layer numbers and training settings, as illustrated in Figure~\ref{fig:performance_plot}. 'Naive Training' denotes that the student model is trained without teacher supervision and replacement training. 'Distill Training' represents that the student model is trained only with teacher supervision. 'CompressTracker' refers to the same training setting in Table~\ref{tab:abl_supervision} $\#8$. It can be observed that as the number of layers increases, there is a corresponding improvement in accuracy. Our CompressTracker shows a noticeable performance boost due to our proposed training strategy, which verifies the effectiveness and generalization ability of our framework.

\textbf{Replacement Training.} 
To evaluate the efficiency and effectiveness of our replacement training strategy, we conduct experiments presented in Table~\ref{tab:ablation_replace}. 'Random' denotes our replacement training, and 'Decouple-300' represents stage-by-stage decoupling. Result of \# 1 aligns with our replacement training with 300 training epochs, while in \# 2, we apply decoupled training, sequentially training and freezing each stage for 75 epochs, followed by 30 epochs of fine-tuning. The 'Decouple-300' (\# 2) approach achieves $64.6\%$ AUC on LaSOT with the same training epochs, marginally lower by $0.6\%$ AUC than our replacement training strategy (\# 1). The 'Decouple-300' approach (\# 2) requires a complex, multi-stage training along with supplementary fine-tuning, which may suffer from suboptimal outcomes at a specific training process. However, our CompressTracker operates on an end-to-end, single-step basis, and can avoid the suboptimal performance issue through its unified training manner, 
which validates the superiority of our replacement training.

\textbf{Progressive Replacement.} In Table~\ref{tab:ablation_replace_ori}, we illustrate the impact of progressive replacement strategy. The first row (\# 1) corresponds to the same setting of CompressTracker, while in the second row (\# 2) we fix the sampling probability as $0.5$ and the student model is trained with 300 epochs followed by 30 finetuning epochs. The absence of progressive replacement leads to a performance degradation of $0.4\%$ AUC, thereby highlighting the efficacy of our progressive replacement approach.

\textbf{Training Epochs.} 
Based on the analysis in Section~\ref{sec:progress}, the optimization steps for each layer are lower than total training steps. Thus, to ensure adequate training of each stage, we increase the training epochs from 300 to 500, and show the result in Table~\ref{tab:abl_epoch}. Extending the training epochs  ensures that student models receive comprehensive training, leading to improved accuracy.

\textbf{Training Time.} 
We compare the training time of our CompressTracker-4 with 500 training epochs, OSTrack, and MixFormerV2-S in Table~\ref{tab:ablation_time}. The training time is recorded on 8 NVIDIA RTX 3090 GPUs. Although our CompressTracker requires a longer training time compared to the OSTrack, the increased computational overhead remains within acceptable limits. Moreover, MixFormerV2-S is trained on 8 Nvidia RTX8000 GPUs, and we estimate this will take roughly 80 hours on 8 NVIDIA RTX 3090 GPUs based on the relative computational capabilities of these GPUs. The training time of our CompressTracker-4 is significantly less than that of MixFormerV2-S, which validate the efficiency and effectiveness of our framework.

\section{Conclusion}
In this paper, we propose a general compression framework, CompressTracker, for visual object tracking. We propose a novel stage division strategy to separate the structural dependencies between the student and teacher models. We propose the replacement training to enhance student's ability to emulate the teacher model. We further introduce the prediction guidance and stage-wise feature mimicking to improve performance. Extensive experiments verify the effectiveness and generalization ability of our CompressTracker. Our CompressTracker is capable of accelerating tracking models while preserving performance to the greatest extent.

{
    \small
    \bibliographystyle{ieeenat_fullname}
    \bibliography{main}
}

\begin{table*}[htp]
  \centering
  \resizebox{\linewidth}{!}{\begin{tabular}{l|ccc|cc|cc|ccc|cc|c}
    \toprule
    \multirow{2}{*}{\textbf{Method}} &
    \multicolumn{3}{c|}{\textbf{LaSOT}} &
    \multicolumn{2}{c|}{\textbf{LaSOT$_{ext}$}} &
    \multicolumn{2}{c|}{\textbf{TNL2K}} &
    \multicolumn{3}{c|}{\textbf{TrackingNet}} &
    \multicolumn{2}{c|}{\textbf{UAV123}} &
    \multirow{2}{*}{\textbf{FPS}} \\
    & AUC & P$_{Norm}$ & P & AUC & P & AUC & P & AUC & P$_{Norm}$ & P & AUC & P & \\
    \midrule
    \textcolor{lightgray}{MixFormerV2-B~\citep{cui2024mixformerv2}} & \textcolor{lightgray}{70.6} & \textcolor{lightgray}{80.8} & \textcolor{lightgray}{76.2} & \textcolor{lightgray}{50.6} & \textcolor{lightgray}{56.9} & \textcolor{lightgray}{57.4} & \textcolor{lightgray}{58.4} & \textcolor{lightgray}{83.4} & \textcolor{lightgray}{88.1} & \textcolor{lightgray}{81.6} & \textcolor{lightgray}{69.9} & \textcolor{lightgray}{92.1} & \textcolor{lightgray}{165} \\
    
    MixFormerV2-S~\citep{cui2024mixformerv2} & 60.6 & 69.9 & 60.4 & 43.6 & 46.2 & 48.3 & 43.0 & 75.8 & 81.1 & 70.4 & 65.8 & 86.8 & 325 \\

    \textbf{CompressTracker-M-S} & \textbf{62.0} $_{\textcolor{darkgray}{88\%}}$ & \textbf{70.9} & \textbf{63.2} & \textbf{44.5} $_{\textcolor{darkgray}{88\%}}$ & \textbf{47.1} & \textbf{50.2} $_{\textcolor{darkgray}{87\%}}$ & \textbf{47.8} & \textbf{77.7} $_{\textcolor{darkgray}{93\%}}$ & \textbf{82.5} & \textbf{73.0} & \textbf{66.9}  $_{\textcolor{darkgray}{96\%}}$ & \textbf{87.1} & \textbf{325} $_{\textcolor{darkgray}{1.97\times}}$ \\

    \bottomrule
    \end{tabular}}
  \caption{\textbf{Compress MixFormerV2.} We compress MixFormerV2 into CompressTracker-M-S with 4 layers, which is the same as MixFormerV2-S including \textit{the dimension of MLP layer}. We report the performance on 5 benchmarks and calculate the performance gap in comparison to the origin MixFormerV2-B. Our CompressTracker-M-S outperforms MixFormerV2-S under the same setting.}
  \label{tab:main_performance_m}
\end{table*}

\begin{table*}[ht]
  \centering
  \resizebox{\linewidth}{!}{\begin{tabular}{l|ccc|cc|cc|ccc|cc|c}
    \toprule
    \multirow{2}{*}{\textbf{Method}} &
    \multicolumn{3}{c|}{\textbf{LaSOT}} &
    \multicolumn{2}{c|}{\textbf{LaSOT$_{ext}$}} &
    \multicolumn{2}{c|}{\textbf{TNL2K}} &
    \multicolumn{3}{c|}{\textbf{TrackingNet}} &
    \multicolumn{2}{c|}{\textbf{UAV123}} &
    \multirow{2}{*}{\textbf{FPS}} \\
    & AUC & P$_{Norm}$ & P & AUC & P & AUC & P & AUC & P$_{Norm}$ & P & AUC & P & \\
    \midrule
    \textcolor{lightgray}{OSTrack-256~\citep{ye2022joint}} & \textcolor{lightgray}{69.1} & \textcolor{lightgray}{78.7} & \textcolor{lightgray}{75.2} & \textcolor{lightgray}{47.4} & \textcolor{lightgray}{53.3} & \textcolor{lightgray}{54.3} & \textcolor{lightgray}{-} & \textcolor{lightgray}{83.1} & \textcolor{lightgray}{87.8} & \textcolor{lightgray}{82.0} & \textcolor{lightgray}{68.3} & \textcolor{lightgray}{-} & \textcolor{lightgray}{105} \\
    
    SMAT~\citep{gopal2024separable} & 61.7 & 71.1 & 64.6 & - & - & - & - & 78.6 & 84.2 & 75.6 & 64.3 & 83.9 & \textbf{158} \\

    \textbf{CompressTracker-SMAT} & \textbf{62.8} $_{\textcolor{darkgray}{91\%}}$ & \textbf{72.2} & \textbf{64.0} & \textbf{43.4} $_{\textcolor{darkgray}{92\%}}$ & \textbf{46.0} & \textbf{49.6} $_{\textcolor{darkgray}{91\%}}$ & \textbf{46.9} & \textbf{79.7} $_{\textcolor{darkgray}{96\%}}$ & \textbf{85.0} & \textbf{75.4} & \textbf{65.9}  $_{\textcolor{darkgray}{96\%}}$ & \textbf{86.4} & 138 $_{\textcolor{darkgray}{1.31\times}}$ \\

    \bottomrule
    \end{tabular}}
  \caption{\textbf{Compress OSTrack for SMAT.} We compress OSTrack into CompressTracker-SAMT with 4 SMAT layers, which is the same as SMAT. We report the performance on 5 benchmarks and calculate the performance gap in comparison to the original OSTrack. Our CompressTracker-SAMT outperforms SMAT under the same setting.}
  \label{tab:main_performance_smat}
\end{table*}

\section{Appendix}

This appendix is structured as follows:
\begin{itemize}
    \item In Appendix~\ref{sec:appendix_gen}, we provide more experiments to verify the strong generalization ability of our CompressTracker.
    \item In Appendix~\ref{sec:appendix_ablation}, we provide more ablation study results.
    \item In Appendix~\ref{sec:appendix_replace}, we show the pseudo code of our CompressTracker.
\end{itemize}

\begin{figure}[htp]
  \centering
  \includegraphics[width=\linewidth]{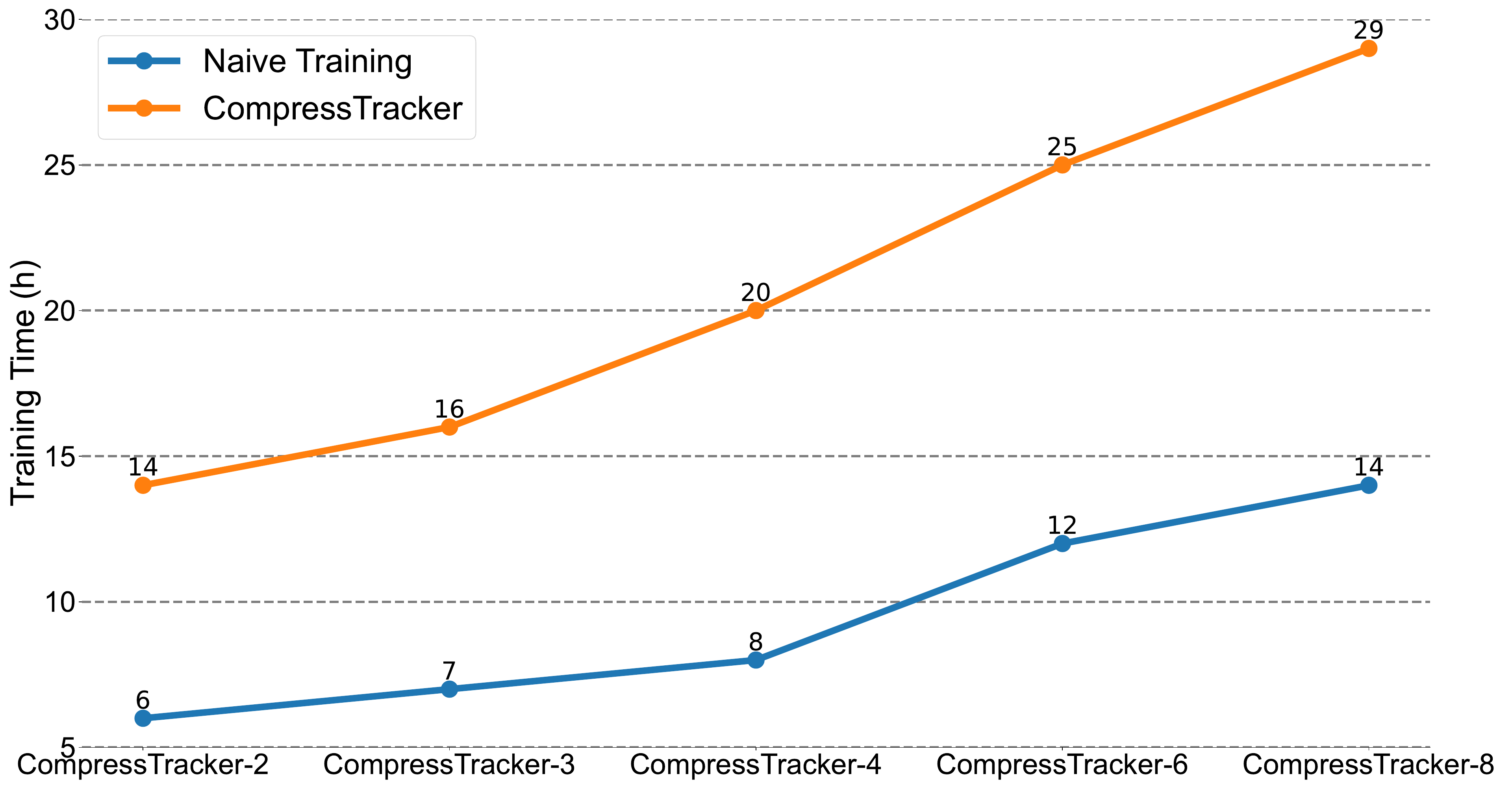}
  \caption{\textbf{Training Time.} }
  \label{fig:plot_time}
\end{figure}

\subsection{Heterogeneous Structure Robustness}
\label{sec:appendix_gen}
\textbf{Compressing MixFormerV2.} To affirm the generalization ability of our approach, we conduct experiments on MixFormerV2~\citep{cui2024mixformerv2} and SMAT~\citep{gopal2024separable}. MixFormerV2-S is a fully transformer tracking model consisting of 4 transformer layers, trained via a complex multi-stages model reduction paradigm. Following MixFormerV2-S, we adopt MixFormerV2-B as teacher and compress it to a student model with 4 layers. The results are shown in Table~\ref{tab:main_performance_m}. Our CompressTracker-M-S share the same structure and channel dimension of MLP layers with MixFormerV2-S and outperforms MixFormerV2-S by about $1.4\%$ AUC on LaSOT.

It's worth noting that although CompressTracker-2 and CompressTracker-M-S have similar inference speeds, MixFormerV2-S and CompressTracker-M-S each contain four transformer layers, whereas CompressTracker-2 only has two. The lower number of transformer layers contributes to the slightly lower performance for CompressTracker-2. Additionally, both CompressTracker-4 and CompressTracker-M-S have four transformer layers, but CompressTracker-M-S has a lower hidden feature dim of MLP layer than CompressTracker-4. As highlighted in MixFormerV2-S~\citep{cui2024mixformerv2}, a reduced feature dimension can lead to decreased accuracy. Consequently, CompressTracker-M-S exhibits slightly lower performance than CompressTracker-4. Moreover, our CompressTracker-4 requires only about 20 hours for training, in contrast to the 120 hours needed for MixFormerV2-S, which also relies on a complex multi-stage training strategy. Besides, the reduction paradigm in MixFormerV2 limits the student model's structure, while our framework supports a diverse range of transformer architectures thanks to our stage division.

\textbf{Compressing OSTrack to SMAT.} SMAT replace the vanilla attention in transformer layer with separated attention. We compress OSTrack into a student model CompressTracker-SMAT, aligning the number and structure of transformer layer with SAMT. We maintain the decoder of OSTrack for CompressTracker-SMAT. CompressTracker-SMAT surpasses SMAT by $1.1\%$ AUC on LaSOT, which demonstrates that our framework is flexible and not limited by the structure of transformer layer. 

Based on results in Table 1 and 2 in main paper and Table~\ref{tab:main_performance_m} and~\ref{tab:main_performance_smat}, our CompressTracker achieves an optimal trade-off between efficiency and performance and is applicable to any \textit{teacher model}, any \textit{resolution}, any \textit{stage number}, any \textit{teacher model size}, and any \textit{student architectures}, which highlights the superiority and strong generalization ability of our CompressTracker.

\subsection{More Ablation Study}
\label{sec:appendix_ablation}

We represent more ablation studies on LaSOT to explore the factors contributing to effectiveness of our CompressTracker. Unless otherwise specified,  teacher model is OSTrack,and student model has 4 encoder layers. The student model is trained for 300 epochs, and the $p_{init}$ is set as $0.5$.

\textbf{Training Time.} 
We compare the training time of CompressTracker with 500 training epochs across different layers in Figure~\ref{fig:plot_time}. 'Naive Training' denotes solely training on groundtruth data with 300 epochs, and 'CompressTracker' represents our proposed training strategy with 500 epochs. The training time is recorded on 8 NVIDIA RTX 3090 GPUs.  Although our CompressTracker requires a longer training time compared to the 'Naive Training', the increased computational overhead remains within acceptable limits.

\subsection{Replacement Training}
\label{sec:appendix_replace}
We present the pseudocode for the training and testing phases of CompressTracker in Algorithm~\ref{alg:comp_train} and Algorithm~\ref{alg:comp_test}, respectively.
Additionally, the pseudocode of OSTrack~\cite{ye2022joint} is also shown in Algorithm~\ref{alg:ostrack}. During training process, we employ Bernoulli sampling to implement a replacement training strategy, while in the test phase, we integrate the student layers and discard the teacher layer.

\begin{algorithm}[h]
\caption{Pseudocode of OSTrack in a PyTorch-like style}
\label{alg:ostrack}
\definecolor{codeblue}{rgb}{0.25,0.5,0.5}
\lstset{
  backgroundcolor=\color{white},
  basicstyle=\fontsize{10pt}{10pt}\ttfamily\selectfont,
  columns=fullflexible,
  breaklines=true,
  captionpos=b,
  commentstyle=\fontsize{7pt}{7pt}\color{codeblue},
  keywordstyle=\fontsize{7pt}{7pt}
}
\begin{lstlisting}[language=python]
# z/x:  RGB image of template/search region
# patch_embed: patch embedding layer, 
# pos_embed_z/pos_embed_z: position embedding for template/search region
# blocks: transformer block layers
# decoder: decoder network

def forward(x, z):
    # patch embedding layer
    x, z = patch_embed(x), patch_embed(z)

    # add position embedding
    x, z = x + pos_embed_x, z + pos_embed_z

    # concat
    x = torch.cat([z, x], dim=1)
    
    # transformer layers
    for i, blk in enumerate(blocks):
        x = blk(x)
    
    # decode the matching result
    x = decoder(x)
\end{lstlisting}
\end{algorithm}

\begin{algorithm}[h]
\caption{Pseudocode of CompressTracker for Training in a PyTorch-like style}
\label{alg:comp_train}
\definecolor{codeblue}{rgb}{0.25,0.5,0.5}
\lstset{
  backgroundcolor=\color{white},
  basicstyle=\fontsize{10pt}{10pt}\ttfamily\selectfont,
  columns=fullflexible,
  breaklines=true,
  captionpos=b,
  commentstyle=\fontsize{7pt}{7pt}\color{codeblue},
  keywordstyle=\fontsize{7pt}{7pt}
}
\begin{lstlisting}[language=python]
# z/x:  RGB image of template/search region
# patch_embed: patch embedding layer, 
# pos_embed_z/pos_embed_z: position embedding for template/search region
# bernoulli_sample: bernoulli sampling function with probability of p
# n_s/n_t: layer number of student/teacher model
# teacher_blocks: transformer block layers of a pretrained teacher
# student_blocks: transformer block layers of student model
# decoder: decoder network

def forward(x, z):
    # patch embedding layer
    x, z = patch_embed(x), patch_embed(z)

    # add position embedding
    x, z = x + pos_embed_x, z + pos_embed_z
    
    # concat
    x = torch.cat([z, x], dim=1)
    
    # replacement sampling
    inference_blocks = []
    for i in range(n):
        if bernoulli_sample() == 1:
            inference_blocks.append(student_blocks[i])
        else:
            for j in range(n_t//n_s):
                inference_blocks.append(teacher_blocks[i*(n_t//n_s) + j])
        
    # randomly replaced transformer layers
    for i, blk in enumerate(inference_blocks):
        x = blk(x)
    
    # decode the matching result
    x = decoder(x)
\end{lstlisting}
\end{algorithm}

\begin{algorithm}[h]
\caption{Pseudocode of CompressTracker for Testing in a PyTorch-like style}
\label{alg:comp_test}
\definecolor{codeblue}{rgb}{0.25,0.5,0.5}
\lstset{
  backgroundcolor=\color{white},
  basicstyle=\fontsize{10pt}{10pt}\ttfamily\selectfont,
  columns=fullflexible,
  breaklines=true,
  captionpos=b,
  commentstyle=\fontsize{7pt}{7pt}\color{codeblue},
  keywordstyle=\fontsize{7pt}{7pt}
}
\begin{lstlisting}[language=python]
# z/x:  RGB image of template/search region
# patch_embed: patch embedding layer, 
# pos_embed_z/pos_embed_z: position embedding for template/search region
# student_blocks: transformer block layers of student model
# decoder: decoder network

def forward(x, z):
    # patch embedding layer
    x, z = patch_embed(x), patch_embed(z)

    # add position embedding
    x, z = x + pos_embed_x, z + pos_embed_z
    
    # concat
    x = torch.cat([z, x], dim=1)

    # transformer layers
    for i, blk in enumerate(student_blocks):
        x = blk(x)
    
    # decode the matching result
    x = decoder(x)
\end{lstlisting}
\end{algorithm}

\end{document}